\documentclass[sigconf, screen]{acmart}
\renewcommand\footnotetextcopyrightpermission[1]{} 
\AtBeginDocument{%
  }

\setcopyright{acmlicensed}
\copyrightyear{2018}
\acmYear{2026}
\acmDOI{XXXXXXX.XXXXXXX}
\acmConference[ACM MM]{Make sure to enter the correct
  conference title from your rights confirmation email}{June 03--05,
  2026}{Brazil}

\usepackage{amsmath,bm}
\usepackage{siunitx}             
\usepackage{graphicx}            
\usepackage{booktabs}            
\usepackage{xcolor}
\usepackage[ruled,linesnumbered,lined,boxed,commentsnumbered]{algorithm2e}
\usepackage{url}

\begin{document}

\title{MS-Mix: Sentiment-Guided Adaptive Augmentation for Multimodal Sentiment Analysis}


\author{Hongyu Zhu}
\affiliation{%
  \institution{Chongqing Institute of Green Intelligent Technology, Chinese Academy of Sciences; Chongqing School, University of Chinese Academy of Sciences}
  \city{Chongqing}
  \country{China}}
\email{zhuhongyu@cigit.ac.cn}

\author{Lin Chen}
\authornotemark[1]
\affiliation{%
  \institution{Chongqing Institute of Green Intelligent Technology, Chinese Academy of Sciences; Chongqing School, University of Chinese Academy of Sciences}
  \city{Chongqing}
  \country{China}}
\email{chenlin@cigit.ac.cn}

\author{Xin Jin}
\affiliation{%
  \institution{School of Engineering,\\ Westlake University}
  \city{Zhejiang}
  \country{China}}
\email{jinxin86@westlake.edu.cn}

\author{Mingsheng Shang}
\authornotemark[1]
\affiliation{%
  \institution{Chongqing Institute of Green Intelligent Technology, Chinese Academy of Sciences; Chongqing School, University of Chinese Academy of Sciences}
  \city{Chongqing}
  \country{China}}
\email{msshang@cigit.ac.cn}

\thanks{$*$ Lin Chen and Mingsheng Shang are the corresponding authors.}

\renewcommand{\shortauthors}{Hongyu Zhu et al.}

\begin{abstract}
Multimodal Sentiment Analysis (MSA) integrates complementary features from text, video, and audio for robust emotion understanding in human interactions. However, models suffer from severe data scarcity and high annotation costs, severely limiting real-world deployment in social media analytics and human-computer systems.
Existing Mixup-based augmentation techniques, when naively applied to MSA, often produce semantically inconsistent samples and amplified label noise by ignoring emotional semantics across modalities.
To address these challenges, we propose MS-Mix, an adaptive emotion-sensitive augmentation framework that automatically optimizes data quality in multimodal settings. Its key components are: (1) Sentiment-aware sample selection strategy that filters incompatible pairs via latent-space semantic similarity to prevent contradictory emotion mixing. (2) Sentiment intensity guided module with multi-head self-attention for computing modality-specific mixing ratios conditioned on emotional salience dynamically. (3) Sentiment alignment loss based on Kullback-Leibler divergence to align predicted sentiment distributions across modalities with ground-truth labels, improving discrimination and consistency.
Extensive experiments on two public datasets with six state-of-the-art backbones confirm that MS-Mix consistently outperforms prior methods, significantly improving robustness and practical applicability for MSA. The source code is available at an anonymous link: https://anonymous.4open.science/r/MS-Mix-review-0C72.
\end{abstract}




\maketitle

\begin{figure}[!ht]
\centerline{\includegraphics[scale=0.56]{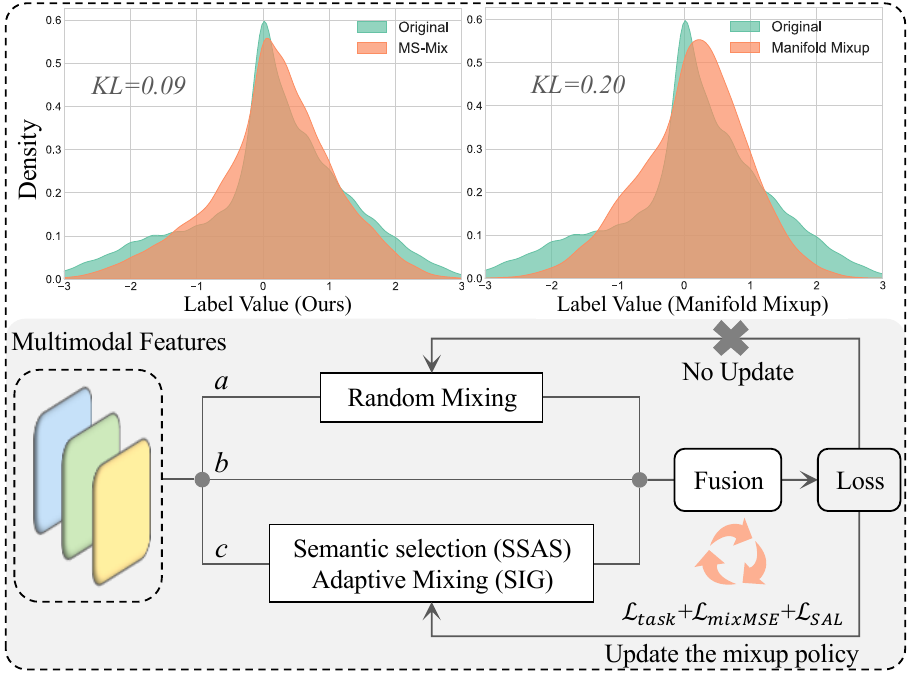}}
    \caption{Top: The comparison of the label distribution generated by MS-Mix and  Manifold Mixup \cite{manifoldmix}. 
    Bottom: a. The traditional methods employ random mixing and an offline optimization. b. Backbone. c. MS-Mix. 
    }
    \vspace{-10pt}
    \label{fig1}
\end{figure}
\section{Introduction}
The perception and understanding of human emotions by Artificial Intelligence (AI) are of great significance for technologies such as human-computer interaction \cite{lv2025diffufuse, huang2026ai}, and multimedia computing \cite{li2019survey}. Multimodal Sentiment Analysis (MSA) has emerged as a critical research frontier in this endeavor \cite{glomo2024, xiao2025exploring}. MSA aims to integrate and interpret complementary emotional features from textual, acoustic, and visual modalities to achieve a robust and accurate understanding of sentiment \cite{wulamu2025enhanced}. By moving beyond unimodal analysis, MSA can capture the complexity of human affective expression, thereby providing a more holistic view of sentiment.

Benefiting from rapid advances in Deep Learning (DL), the field of MSA has witnessed substantial progress in developing model architectures \cite{Mult2019, glomo2024}.
Techniques such as cross-modal transformers \cite{Mult2019} and advanced fusion mechanisms \cite{glomo2024} have demonstrated remarkable capability in modeling inter-modal interactions and extracting discriminative features. 
Nevertheless, the performance of data-driven DL models remains fundamentally constrained by the scale and quality of annotated data \cite{vanilla_mixup,borovits2026addressing}. This data scarcity often results in overfitting and limited generalization ability \cite{2014dropout,weldy2026simulated}. Furthermore, constructing MSA datasets is costly, as it requires collecting reliable, human-annotated multimodal sentiment data. 
To mitigate data scarcity, the mixup strategy was introduced to generate virtual training samples by convexly interpolating between original data points and their corresponding labels \cite{vanilla_mixup}. However, existing offline methods, including general domain techniques such as CutMix \cite{2019cutmix} and SaliencyMix \cite{2021saliencymix}, as well as multimodal specific approaches like $\mathcal{P}ow$Mix \cite{powmix} and MultiMix \cite{multimix}, often depend on randomly paired samples and offline optimization strategies, thereby neglecting the underlying emotional semantics. This may result in the blending of semantically inconsistent samples. Moreover, these methods typically use fixed or uniformly distributed mixing ratios that do not adapt to varying emotional intensities across modalities. As shown in Fig.~\ref{fig1}, MS-Mix generates augmented label distributions that align significantly closer to the ground truth than traditional mixup methods represented by Manifold Mixup, achieving lower Kullback-Leibler (KL) divergence.

To address these limitations, we propose MS-Mix. Unlike prior methods that rely on random mixing or static ratios, MS-Mix introduces the first fully sentiment-guided adaptive augmentation framework to ensure semantically consistent and high-quality sample generation. As shown in Fig.~\ref{fig2}, our method includes three key innovations: jointly optimizing sample selection, mixing weights, and cross-modal distribution alignment.
Experimental results on real-world MSA datasets show the obvious advantages of MS-Mix. Specifically, on the MOSI \cite{MOSI} and MOSEI \cite{MOSEI} datasets, MS-Mix achieves consistent improvements on all six backbones, with the comprehensive metric exceeding all baselines by an average of 1.29\% on MOSI and 1.16\% on MOSEI, respectively.

The main contributions are summarized as follows:
\begin{itemize}
    \item We design a SASS strategy that leverages semantic similarity in the latent space to filter out incompatible sample pairs, effectively preventing mixtures of samples with contradictory emotions.
    \item We introduce a SIG mixing module, implemented with multi-head self-attention, to dynamically determine modality-specific mixing ratios based on emotional salience.
    \item We propose SAL, a KL divergence-based regularization that aligns predicted sentiment distributions with ground-truth labels, enhancing the model’s generalization capability.
\end{itemize}


\section{Related Work}
\subsection{Multimodal Sentiment Analysis}
Prior research in MSA has predominantly focused on improving two main paradigms: feature-fusion strategy-centric \cite{TFN2017, LMF2018, 2018memoryzadeh, Mult2019, 2021progressivelv, glomo2024} and feature-encoder method-centric \cite{2021attentionliang, yang2022dis, lin2023multi, ALMT2023} approaches.

Fusion strategy-centric approaches in MSA primarily focused on designing effective fusion strategies to integrate features from different modalities. Zadeh et al. \cite{TFN2017} introduced the Tensor Fusion Network (TFN), which explicitly models intermodal interactions via the 3-fold Cartesian product. Building on this work, Z. Liu et al. \cite{LMF2018} proposed Low-rank Multimodal Fusion (LMF), an efficient approach that improves computational performance through parallel tensor and weight decomposition. MuIT \cite{Mult2019} incorporates cross-modal attention and achieves multimodal fusion across unaligned datasets.
A recent advancement in MSA is the attempt to decouple the features into shared and unique information. For instance, Li et al. \cite{li2023decoupled} proposed a Decoupled Multimodal Distillation (DMD) method that distills cross-modal knowledge in decoupled feature spaces and alleviates the inherent multimodal heterogeneity. The Global Local Modal (GLoMo) \cite{glomo2024} Fusion framework integrates multiple local representations within each modality and effectively combines local and global information. 

In feature encoder-centric approaches, the primary objective is to improve the feature representation for each modality. Yang et al. \cite{yang2022dis} employ adversarial learning to model both modality-invariant and modality-specific subspaces within multimodal fusion.
Han et al. \cite{han2021improving} enhance multimodal fusion performance through hierarchical mutual information maximization.
To address the challenge of multimodal heterogeneity in MSA, Li et al. \cite{lin2023multi} conceptualized the learning process for each modality as a set of sub-tasks, thereby effectively reducing discrepancies between modalities and enhancing representation consistency. Furthermore, to suppress sentiment-irrelevant and conflicting information across modalities, in recent work, Zhang et al. \cite{ALMT2023} proposed the Adaptive Language-guided Multimodal Transformer (ALMT), which learns a unified hypermodality representation guided by language at multiple scales.

Despite these advancements, MSA systems remain constrained by the scarcity and high annotation costs of multimodal data, which often lead to overfitting and limited generalization \cite{multimix, powmix}. This highlights the need for effective data augmentation techniques tailored to multimodal emotional data.

\subsection{Mixup-based Augmentation}
Mixup and its variants have been widely adopted in DL \cite{puzzlemix, automix, SUMIX2024, jin2024survey, fu2025advmixup}. The original Mixup method \cite{vanilla_mixup} operates in the input space by generating virtual samples through linear interpolation between two randomly selected data points and their corresponding labels.
This simple yet effective strategy encourages models to behave linearly across classes. Subsequent extensions, such as Manifold Mixup \cite{manifoldmix}, generalize the interpolation operation to hidden representations, further enhancing the smoothness of decision boundaries. 
In the latest work, AdvMixUp \cite{fu2025advmixup} introduces a sample-dependent, feature-level interpolation mask generated by a compact Mask Generator, optimized via adversarial training to produce challenging "hard" mixed samples near decision boundaries.
Recent state-of-the-art approaches \cite{automix, SUMIX2024, fu2025advmixup} adaptively generate mixed samples using learnable mixing ratios and feature representations in an end-to-end manner.

In the multimodal domain, several works have extended Mixup to leverage cross-modal interactions. For instance, MultiMix \cite{multimix} performs independent Mixup operations within each modality. VLMixer \cite{2022vlmixer} integrates cross-modal Mixup with contrastive learning to convert unimodal text inputs into multimodal representations. To enhance robustness to missing modalities, $M^3$ixup \cite{lin2024adapt} extends mixup to both the representation and contrastive loss levels to improve cross-modal alignment and the capture of dynamics. Recently, building on MultiMix, $\mathcal{P}$owMix \cite{powmix} introduced a weight-aware mixing strategy that dynamically modulates the mixing coefficients based on the estimated importance of each modality.

However, most existing methods rely on random sampling and offline mixing, which overlook the semantic structure and emotional coherence of multimodal data. This often results in semantically inconsistent mixtures, such as mixing samples with opposing emotions, which introduces label noise. These limitations underscore the need for more semantically aware and adaptively controlled mixing strategies tailored to MSA. 
Therefore, we propose MS-Mix, a novel sentiment-guided adaptive augmentation framework that effectively avoids the mixing of contradictory emotions and adaptively generates discriminative multimodal emotional features. 

\begin{figure*}[htbp]
\centerline{\includegraphics[scale=0.53]{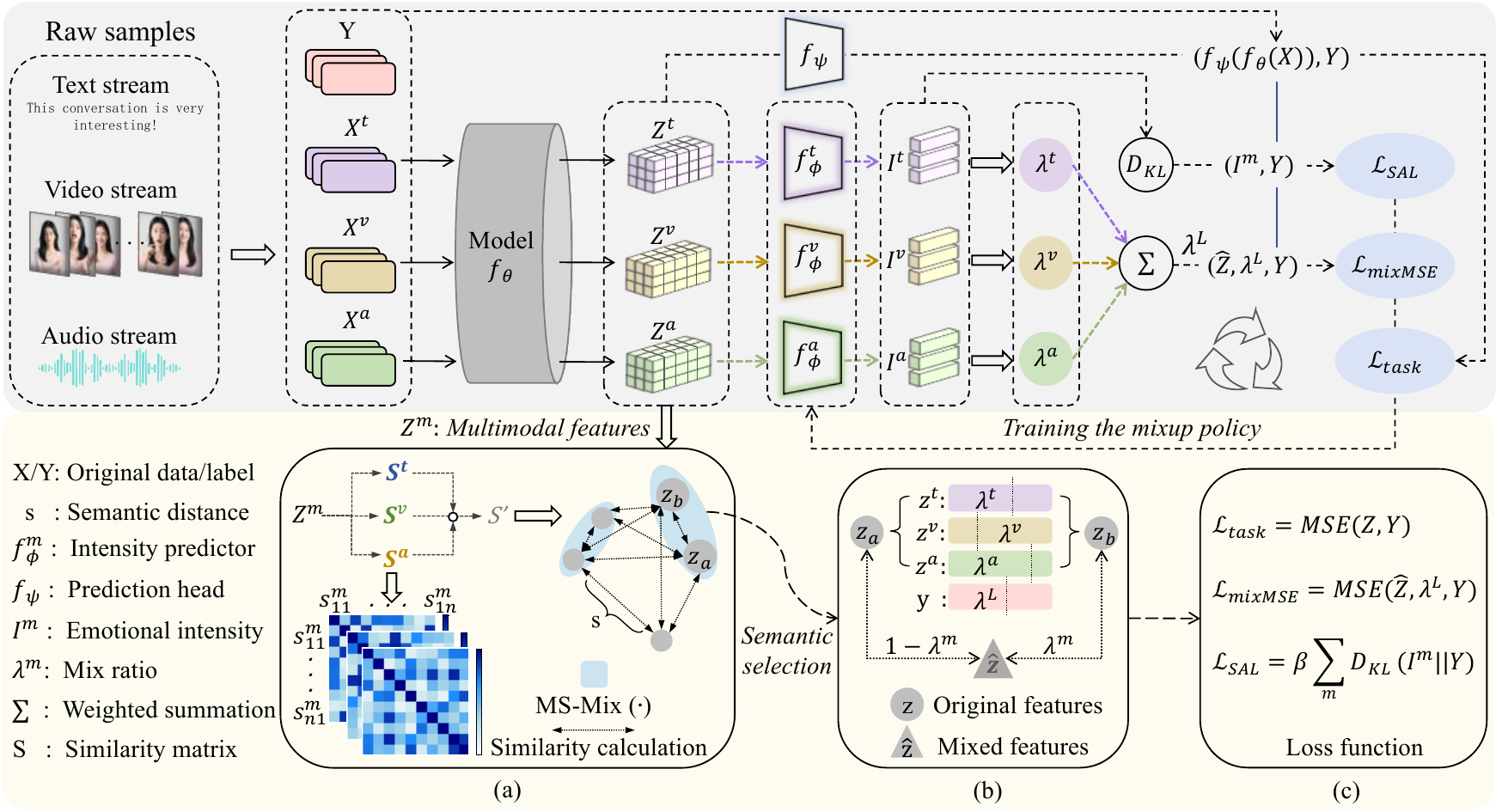}}
    \caption{An overview of the MS-Mix framework. (a) SASS strategy, (b) SIG module, and (c) SAL ($\mathcal{L}_{SAL}$) regularization term.
    }
    \vspace{-5pt}
    \label{fig2}
\end{figure*}

\section{Methodology}
In this section, we first introduce the task definition and the overview of the MS-Mix framework. Then we describe the details of each module. The pseudocode for MS-Mix is presented in Appendix A.1.

\subsection{Task Definition and Framework Overview}
Given a MSA dataset with $n$ samples, ${\mathbf{X}=(\mathbf{x}_1, \mathbf{x}_2, ...,\mathbf{x}_n)}$ and ${\mathbf{Y}=}$ ${(y_1, y_2, ...,y_n)}$ where $y_i$ are continuous sentiment intensity scores, each $\mathbf{x}_i \in \mathbf{X}$ comprises data from the text ($t$), video ($v$), and audio ($a$) modalities, denoted as $\{\mathbf{x}_i^t, \mathbf{x}_i^v, \mathbf{x}_i^a\}$. The goal of the MSA task is to learn a mapping function $\mathcal{F}: \mathbf{X} \mapsto \mathbf{Y}$ to predict the continuous intensity of each emotion category. This mapping is realized by a backbone $f_\theta$ followed by a prediction head $f_\psi$, using a network $f_\psi\circ f_\theta$. The network parameters $\psi$ and $\theta$ are typically optimized by minimizing a task-specific loss $\mathcal{L}_{task}$:
\begin{equation}
    \begin{aligned}
    \label{eq:1}
        \min_{\psi, \theta} \mathcal{L}_{task} (f_\psi(f_\theta(\mathbf{x})), y).
    \end{aligned}
\end{equation}

MS-Mix is a latent-space data augmentation method that operates on the multimodal features output by the encoder and is applied before feature fusion. As illustrated in Fig.\ref{fig2}, the input to MS-Mix consists of multimodal features derived from the model’s encoder outputs. The SASS strategy first identifies feature pairs suitable for mixing, which are then adaptively blended through the SIG module to generate augmented features. Both the original and synthesized features are subsequently fed into the fusion module. Additionally, the SAL serves as an auxiliary regularization term to jointly optimize the entire network and the SIG module. 

As a common starting point for many mixup-based augmentation methods, vanilla mixup \cite{vanilla_mixup} employs a ratio $\lambda$ to construct a mixed sample ($\hat{\mathbf{x}}, \hat{y}$) by performing global linear interpolation directly on the sample pair $\{(\mathbf{x}_i, y_i), (\mathbf{x}_j, y_j)\}$: 
\begin{equation}
    \begin{aligned}
    \label{eq:2}
        \hat{\mathbf{x}} = \lambda \cdot \mathbf{x}_i + (1-\lambda) \cdot \mathbf{x}_j, \\
        \hat{y} = \lambda \cdot y_i + (1-\lambda) \cdot y_j,
    \end{aligned}        
\end{equation}
where the $\lambda$ is sampled from $Beta(\alpha,\alpha)$ distribution. We define the process of Eq.\ref{eq:1} as follows: Given the sample mixup function $\mathrm{H}(\cdot)$, the label mixup function $\mathrm{G}(\cdot)$, and a mixing ratio $\lambda$, we can generate the mixed sample ($\hat{\mathbf{x}}, \hat{y}$) with $\hat{\mathbf{x}} = \mathrm{H}(\mathbf{x}_i, \mathbf{x}_j, \lambda)$, $\hat{y} = \mathrm{G}(y_i, y_j, \lambda)$.

Mixing of original data can easily lead to sentiment semantic confusion \cite{SUMIX2024}. Therefore, for each batch of data, we choose to learn different mixing ratios $\lambda^m$ to mix the features $\mathbf{Z}^m=f_\theta(\mathbf{X})\in \mathbb{R}^{B \times d^m}$ of each modality in the latent space separately (Eq. \ref{eq:3}), and calculate the label ratio $\lambda^L$ to mix the labels as follows:

\begin{equation}
    \begin{aligned}
    \label{eq:3}
        \hat{\mathbf{z}}^m = \mathrm{H}(f_\theta(\mathbf{x}_i^m), f_\theta(\mathbf{x}_j^m), \lambda^m), \mathbf{x}^m \in \mathbf{X}, \\
        \hat{y} = \mathrm{G}((y_i, y_j), \lambda^L), y \in \mathbf{Y},
    \end{aligned}
\end{equation}
where $\hat{\mathbf{z}}^m$ represents the mixed features for modality $m \in \mathcal{M}$, and $\mathcal{M} = \{t, v, a\}$ represents the set of all data modalities. $B$ is the mini-batch size, $d^m$ is the hidden space dimension of modality $m$, and $f_\theta$ is the backbone. 

Finally, the mixed and original features will be fused into a set for subsequent multimodal feature fusion.

\begin{figure*}[htbp]
\centerline{\includegraphics[scale=1.05]{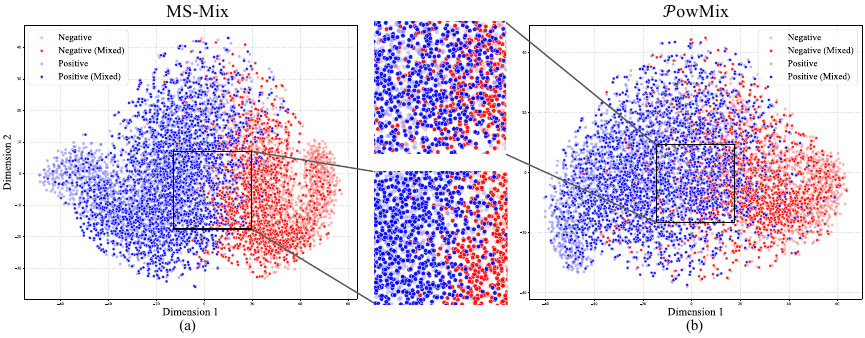}}
    \caption{The t-SNE visualization of the original features and mixed features generated by MS-Mix (a) and $\mathcal{P}$owMix (b) on the MOSEI dataset using the MISA model. We employ a color scheme (blue/red) to differentiate the positive and negative categories, and use transparency to distinguish the original features from the mixed ones.
    }
    \vspace{-5pt}
    \label{fig_sne}
\end{figure*}

\subsection{Sentiment-Aware Sample Selection}
Emotional expressions exhibit significant variation across samples. While random mixing can enhance feature smoothness, it may produce semantically inconsistent samples by blending instances with opposing emotions, leading to neutral labels and confused features. It's a critical factor that previous methods overlooked \cite{powmix}. To address this, we propose the SASS strategy, which screens feature pairs within each batch based on emotional semantic similarity before mixing.
Specifically, we compute cosine similarity to measure semantic relatedness. We first perform $L_2$ normalization on the features of each modality $\mathbf{Z}^m$ to ensure consistent feature scaling:



\begin{equation}
    \begin{aligned}
    \label{eq:4}
    \mathbf{z}^m_{i,\text{norm}} 
    = \frac{\mathbf{z}^m_i}{\sqrt{\sum_{k=1}^{d^m} \left( z^m_{i,k} \right)^2}},\quad i=1,\dots,B,
    \end{aligned}
\end{equation}
where \(\mathbf{Z}^m \in \mathbb{R}^{B \times d^m}\) stacks the feature vectors \(\mathbf{z}^m_i\).

The cross-modal similarity matrix is then obtained as:

\begin{equation}
    \begin{aligned}
    \label{eq:5}
    \mathbf{S} = \frac{1}{|\mathcal{M}|}\sum_{m \in \mathcal{M}} \mathbf{Z}^m_{\text{norm}} (\mathbf{Z}^m_{\text{norm}})^\top.
    \end{aligned}
\end{equation}

Since \(\mathbf{S}\) is symmetric, we consider only its upper triangular part (excluding the diagonal). From all pairs \(\{(\mathbf{z}_i, y_i), (\mathbf{z}_j, y_j)\}\) with \(i < j\) satisfying \(s_{ij} > \delta = 0.2\), we randomly select \(B\) pairs for mixing. A suitable threshold \(\delta\) effectively prevents semantic confusion in the mixed samples while still providing sufficient data diversity. If the number of pairs is less than $B$, we fall back to selecting the $B$ pairs with the highest similarity from the upper triangular part of $\mathbf{S}$.



\subsection{Sentiment Intensity Guided Mixing Module}
The mixing ratio $\lambda$ is one of the most important hyperparameters in mixup-based methods, controlling the degree of mixing between two or more samples. While $\lambda$ is usually sampled from the $Beta(\alpha, \alpha)$ distribution, recent works \cite{puzzlemix, automix, fu2025advmixup} have shown that dynamically optimizing \(\lambda\) based on sample properties improves alignment between mixed data and labels.

To ensure that samples with richer emotional semantics contribute more substantially to the mixing process, we propose the SIG mixing module. Specifically, for each modality, a Multi-Head self-Attention (MHA) \cite{attention} encoder is trained to predict emotional intensity values $\mathbf{I}^m$. This encoder comprises an MHA layer, a residual connection, layer normalization, and a tanh activation function. The modality-specific emotional intensity predictor is denoted as $f_\phi^m$ (Eq. \ref{MHA1}-\ref{MHA3}).
These predictions are then used as weights to dynamically adjust the mixing ratios $\lambda^m$ during augmentation. 
\begin{equation}
    \begin{aligned}
    \label{MHA1}
        \mathbf{h}_i=\mathrm{Softmax}\left ( \frac{\mathbf{Z}^m\mathbf{W}_i^Q \mathbf{Z}^m({\mathbf{W}_i^K})^\top}{\sqrt{d_{k}}}\right )\mathbf{Z}^m\mathbf{W}_i^V, \\
    \end{aligned}
\end{equation}

\begin{equation}
    \begin{aligned}
    \label{MHA2}
        \mathrm{MHA}(\mathbf{Z}^m) & = \mathrm{LN}(\mathrm{Concat}(\mathbf{h}_1, \mathbf{h}_2,...,\mathbf{h}_h)\mathbf{W}^O+\mathbf{Z}^m),
    \end{aligned}
\end{equation}

\begin{equation}
    \begin{aligned}
    \label{MHA3}
        \mathbf{I}^m= f_\phi^m(\mathbf{Z}^m) = \mathrm{tanh}(\mathrm{GlobalPool}(\mathrm{MHA}(\mathbf{Z}^m))). \\
    \end{aligned}
\end{equation}
where $\mathbf{W}^{\{Q, K, V\}}={\{\mathbf{W}^Q,\mathbf{W}^K,\mathbf{W}^V\}}$ are learnable weight matrices that map the input features to queries $Q$, keys $K$, and values $V$, $d_k$ is the dimension of $K$, $LN$ denotes Layer Normalization \cite{ba2016layer}, the number of attention heads $h$ is set to 4, and $\mathbf{W}^O$ is the output linear transformation matrix. 

Based on the intensity of emotions $\mathbf{I}^m={(I^m_1, I^m_2, ..., I^m_B)}$ in Eq. \ref{MHA3}, we perform Min-Max normalization on it within each modality, which can further calculate the intermediate mixing weights ${\bm\omega}^m_i$ for each feature pair $\{ (\mathbf{z}_i, y_i),(\mathbf{z}_j, y_j)\}$:

\begin{equation}
    \begin{aligned}
    \label{eq:10}
        \bm\omega_i^m = \frac{|I_i^m|-min(\mathbf{I}^m)}{max(\mathbf{I}^m)-min(\mathbf{I}^m)+\epsilon}, \\
    \end{aligned}
\end{equation}
where $\epsilon$ represents the minimum value that prevents division by zero and is set to $10^{-8}$. Due to the convex combination strategy, the adaptive mixing ratio $\lambda^m_{i,j}$ between the $\mathbf{z}_i$ and $\mathbf{z}_j$ is:
\begin{equation}
    \begin{aligned}
    \label{eq:11}
        \lambda^m_{i,j}= (\frac{\bm\omega_i^m}{\bm\omega_i^m+\bm\omega_j^m} + \lambda_{base})/2,
    \end{aligned}
\end{equation}
where the $\lambda_{base}$ sampled from the $\text{Beta}(\alpha, \alpha)$. Then,
we calculate the average of the mixing ratios $\lambda^m$ for each modality to obtain the mixing ratio $\lambda^L_{i,j}$ of the labels:
\begin{equation}
    \begin{aligned}
    \label{eq:12}
        \lambda^L_{i,j} =\frac{1}{|\mathcal{M}|} \sum_{m \in \mathcal{M}}\lambda_{i,j}^m. \\
    \end{aligned}
\end{equation}

Finally, we can get the mixed feature $\hat{\mathbf{z}}^m$ and mixed label $\hat{y}$ using Eq. \ref{eq:3}. Unlike previous works \cite{vanilla_mixup, manifoldmix, 2019cutmix, 2021saliencymix}, our proposed SIG module adaptively determines the mixing ratios between samples based on sentiment intensity, thereby maintaining end-to-end trainability and enabling continuous optimization throughout training.

\subsection{Sentiment Alignment Loss Function} 
To enhance the accuracy of the emotion intensity predictor, we introduce the SAL as an additional regularization term to align the predicted distribution with the ground-truth labels. 
Since the emotional labels are continuous values \cite{MOSI, MOSEI}, we consider the mixup regression task. Therefore, we get the mapping between the mixed $\hat{\mathbf{x}}$ and $\hat{y}$ by optimizing mixed Mean Squared Error (MSE) loss $\mathcal{L}_{mixMSE}$ (Eq.\ref{eq:13}). By minimizing $\mathcal{L}_{mixMSE}$ (Eq.\ref{eq:14}), the consistency between the mixed samples and labels is constrained.

\begin{equation}
    \begin{aligned}
    \label{eq:13}
        \mathcal{L}_{mixMSE} & = \lambda^L_{i,j} \cdot MSE(f_\psi (f_\theta(\hat{\mathbf{x}_i})),y_i) \\
        &+(1-\lambda^L_{i,j}) \cdot MSE(f_\psi (f_\theta(\hat{\mathbf{x}_j})),y_j),
    \end{aligned}
\end{equation}

\begin{equation}
    \begin{aligned}
    \label{eq:14}
        \min_{\psi,\theta} \mathcal{L}_{mixMSE} (f_\psi (f_\theta(\hat{\mathbf{x}})), \hat{y}).
    \end{aligned}
\end{equation}

The KL divergence \cite{KLloss} is commonly used to measure the amount of information lost when an approximate distribution is used to represent a true distribution. A smaller KL divergence indicates that the two distributions are more similar. Thus, in this study, we introduce a SAL based on the KL divergence to align the predicted sentiment distribution with the ground-truth labels. 

Specifically, we convert $\mathbf{I}^m$ and ground truth $\mathbf{Y}$ into probability distributions $\mathbf{P}^m$ and $\mathbf{P}^L$ via softmax:
\begin{equation}
    \begin{aligned}
    \label{eq:div}
        P^m_i=\frac{exp(I^m_i)}{\sum^B_{j=1}exp(I^m_j)}, P^L_i=\frac{exp(y_i)}{\sum^B_{j=1}exp(y_j)}. 
    \end{aligned}
\end{equation}

The KL divergence between them is:

\begin{equation}
    \begin{aligned}
    \label{eq:kl}
    \mathrm{KL}(\mathbf{P}^L||\mathbf{P}^m)=\frac{1}{B}\sum^B_{j=1}P^L_j(logP^L_j-logP^m_j).
    \end{aligned}
\end{equation}

We scale this term and sum over all modalities: 
\begin{equation}
    \begin{aligned}
    \label{eq:kl_loss}
        \mathcal{L}_{SAL}=\sum_{m\in{\{t, v, a\}}}\mathrm{KL}(\mathbf{P}^L||\mathbf{P}^m) \cdot \beta,
    \end{aligned}
\end{equation}
where $\beta$ denotes the scaling factor and is set to $10^3$. 

\begin{figure*}[htbp]
\centerline{\includegraphics[scale=1.00]{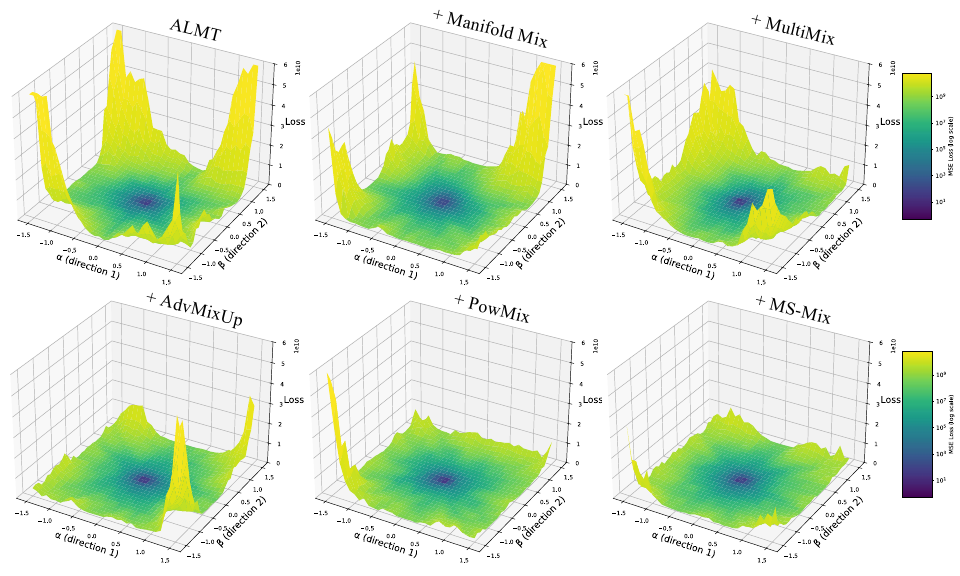}}
    \caption{Visualization of loss landscapes comparing Mixup variants and MS-Mix on MOSEI using ALMT.
    }
    \vspace{-5pt}
    \label{fig_loss}
\end{figure*}

In summary, the total loss $\mathcal{L}_{total}$ for model optimization is expressed as a joint loss defined as follows: 
\begin{equation}
    \begin{aligned}
    \label{eq:loss}
        \mathcal{L}_{total}=\mathcal{L}_{task} + \xi_1 \cdot \mathcal{L}_{mixMSE} + \xi_2 \cdot \mathcal{L}_{SAL}.
    \end{aligned}
\end{equation}
where $\xi_1$ and $\xi_2$ are used to regulate mixed MSE loss and SAL loss.
By enforcing strict cross-modal distribution alignment, SAL can act as a robust regularizer that steers the optimization trajectory toward flatter minima. Consequently, it substantially enhances the model's generalization capability, an advantage empirically corroborated by the loss landscape depicted in Fig. \ref{fig_loss}.

\section{Experiments}
This section presents a series of experiments to evaluate our method.
The comparative results are summarized in Tables \ref{table_mosi},\ref{table_mosei}. Extensive ablation experiments were conducted in Table \ref{table_ablation}, as well as visualizations of the model performance in Fig. \ref{fig_sne} and \ref{fig_loss}. The \textbf{Bold} indicates the best performance, and $\dag$ represents the result reported in \cite{MSENA2022}.

All hyperparameter analysis experiments, additional comparative tests, ablation experiments, and algorithm complexity analysis are detailed in Appendix C.1-C.4, respectively.

\subsection{Datasets}
To comprehensively evaluate our method, we conduct experiments on the widely-used MSA benchmarks: CMU-MOSI (MOSI)~\cite{MOSI} and CMU-MOSEI (MOSEI)~\cite{MOSEI}. 
The MOSI~\cite{MOSI} dataset consists of 2,199 English movie review segments. The MOSEI \cite{MOSEI} dataset is a larger benchmark containing 23,453 segments from more than 1,000 speakers. Both datasets are annotated with sentiment intensity scores ranging from -3 (strongly negative) to +3 (strongly positive). Comprehensive dataset descriptions are available in Appendix A.2. 


\subsection{Backbones and Baseline Methods}
We compared the proposed method against four representative approaches: Manifold Mixup (Manifold Mix)~\cite{manifoldmix}, MultiMix~\cite{multimix}, $\mathcal{P}$owMix~\cite{powmix}, and AdvMixUp~\cite{fu2025advmixup}. These methods are selected as baselines because they collectively represent the major design paradigms in mixup-based data augmentation. They range from classic latent-space interpolation~\cite{manifoldmix} and modality-specific mixing designed for multimodal fusion~\cite{multimix}, to dynamic importance weighting~\cite{powmix} and adversarial sample generation~\cite{fu2025advmixup}. 

These comparisons were conducted across six widely-used MSA architectures: TFN~\cite{TFN2017}, LMF~\cite{LMF2018}, MuIT~\cite{Mult2019}, MISA~\cite{MISA2020}, ALMT~\cite{ALMT2023}, and GLoMo~\cite{glomo2024}.
These backbones encompass a broad range of feature extraction paradigms and fusion strategies, spanning from early influential works to more recent state-of-the-art approaches in MSA. All models were implemented using available code and evaluated under identical settings using the M-SENA framework~\cite{MSENA2022}.

\subsection{Experimental Settings}
\paragraph{Evaluation Metrics}
We use the metrics defined in the M-SENA unified framework \cite{MSENA2022} to ensure comparable results. These include binary accuracy ($ACC_2$) and F1-score (\textit{F1}), multiclass accuracy ($ACC_5$, $ACC_7$) and Mean Absolute Error (MAE). The prefix "\textit{w-}" denotes the absence of neutral-labeled samples. Furthermore, to facilitate intuitive and comprehensive comparisons, we have presented the average values (\textit{Avg.}) for all \textit{ACC} and \textit{F1} metrics.

\paragraph{Experimental Details}
All experiments are conducted on NVIDIA GeForce RTX 3090 Ti and are conducted five times with different random seeds. In MS-Mix, the base mixing ratio $\lambda_{base}$ is sampled from a symmetric Beta distribution with $\alpha = 2.0$, and the similarity threshold $\delta$= 0.2. The loss weights $\xi_1$ and $\xi_2$ are set to 0.7 and 0.5, respectively. Additionally, the number of attention heads $\mathbf{h}$= 4. 
\begin{table*}[!ht]
\centering
\caption{Results of Various Approaches on The \textbf{MOSI} Dataset (Mean ± std over 5 Independent Runs).}
 
\resizebox{0.9\linewidth}{!}{

\begin{tabular}{c | c c c c  c c c | c }
\toprule
\textbf{MODEL}  & w-ACC$_2$(\%)$\uparrow$& ACC$_2$(\%)$\uparrow$ & w-F1(\%)$\uparrow$ & F1(\%)$\uparrow$ & ACC$_5$(\%)$\uparrow$ & ACC$_7$(\%)$\uparrow$ & MAE $\downarrow$ & Avg.(\%)$\uparrow$ \\
\hline
\textbf{TFN}$^\dag$ & 79.08 & 77.99 & 79.11 & 77.95 & 39.39 & 34.46 & 0.947 & 64.66	 \\

+ Manifold Mix & 79.26±1.07 & 78.13±1.12 & 79.07±1.04 & 78.15±1.11 & 39.02±1.15 &  35.23±0.74 & 0.921±0.012 & 64.81±0.95 \\

+ MultiMix & 79.43±0.63 & 78.22±1.51 & 79.31±1.05 & 78.03±1.12 &  39.48±0.71 & 34.95±0.26 & 0.925±0.013 & 64.90±0.57  \\

+ AdvMixUp &  78.48±2.56  & 78.13±0.38  & 79.50±1.17  & 77.74±1.41  &  39.58±0.70  &  34.98±0.63 &  0.923±0.006 &  64.73±0.51 \\

+ $\mathcal{P}ow$Mix & 79.64±1.42 & 78.18±0.83 & 79.35±0.96 & 77.95±0.97 &  40.11±1.08 & \textbf{36.16}±0.44 &  0.919±0.006 & 65.23±0.31 \\

+ MS-Mix (ours) & \textbf{80.22}±0.55 & \textbf{78.72}±0.55 & \textbf{79.99}±1.11 &  \textbf{78.39}±0.67 & \textbf{40.45}±0.68 & 35.89±0.20 & \textbf{0.914}±0.007 & \textbf{65.61}±0.29 \\
\hline


\textbf{LMF}$^\dag$ & 79.18 & 77.90 & 79.15 & 77.80 & 38.13 &  33.82 & 0.950 & 64.33\\
 
+ Manifold Mix & 79.23±1.12 &  77.85±1.28 & 79.34±0.94 & 77.95±0.68 & 37.84±1.22 & 34.17±0.32 & 0.943±0.009 & 64.40±0.21\\

+ MultiMix &  79.92±0.81  & 78.19±0.24  & 80.29±0.89   & 78.58±0.46  & 38.51±0.49   & 35.80±0.65  & 0.925±0.006 & 65.21±0.31  \\

+ AdvMixUp & 79.27±1.06 & 77.84±0.25 &  79.40±0.62 & 78.63±0.53 & 38.16±0.48 & 34.70±0.90  &  0.928±0.004 & 64.67±0.41 \\

+ $\mathcal{P}ow$Mix &  80.05±0.45 &  78.90±0.30 &  79.86±0.81 &  78.55±0.54 & 38.95±0.83 & 35.91±0.51 &  0.915±0.009 &  65.37±0.39\\

+ MS-Mix (ours) & \textbf{82.02}±0.26 & \textbf{79.16}±0.32 & \textbf{82.01}±0.16 & \textbf{79.27}±0.68 & \textbf{41.90}±0.42 &  \textbf{36.52}±0.50 & \textbf{0.879}±0.003 & \textbf{66.81}±0.13 \\
\hline


\textbf{MuIT}$^\dag$ & 80.98 & 79.71 & 80.95 & 79.63 & 42.68 & 36.91 & 0.878 & 66.81 \\
 
+ Manifold Mix & 81.01±0.58 & \textbf{80.38}±1.26 & 80.04±1.95 & 79.34±0.28 & 41.14±1.67 & 36.66±1.37 & 0.864±0.019 & 66.43±0.83 \\

+ MultiMix & 81.49±0.72 & 80.20±0.67 & 81.62±0.48 & \textbf{80.19}±0.91 & 42.76±0.87 & 37.37±0.69 & 0.856±0.004  & 67.27±0.47 \\

+ AdvMixUp & 81.20±0.55 &  79.19±0.27 & 81.64±0.39 & 79.57±0.74 &  42.64±0.75 & 36.51±0.46 & 0.859±0.007  & 66.79±0.28 \\


+ $\mathcal{P}ow$Mix & 81.44±0.56 & 79.70±0.62 & 81.38±0.52 & 79.53±0.84 & 41.57±0.62 & 36.39±0.87 & 0.876±0.04 & 66.67±0.27\\

+ MS-Mix (ours) & \textbf{82.02}±0.28 &  79.69±0.66 & \textbf{81.97}±0.40 &  79.89±0.84 & \textbf{43.33}±0.67 & \textbf{37.48}±0.66 & \textbf{0.830}±0.006 & \textbf{67.40}±0.11   \\
\hline


\textbf{MISA}$^\dag$ & 83.54 & 81.84 & 83.58 & 81.82 & 47.08 &  41.37 & 0.777 & 69.87 \\
 
+ Manifold Mix &  83.03±0.84 &  81.49±0.53 &  83.07±0.84 &  81.47±0.51 &  48.00±0.85 &  41.84±0.38 &  0.762±0.013  & 69.82±0.42 \\

+ MultiMix &  82.91±0.66 &  81.80±0.41 & 82.97±0.91 &  81.88±0.51 & 47.40±0.77 & 42.04±0.36 & 0.751±0.005 & 69.83±0.31 \\

+ AdvMixUp &  83.44±0.47 & 81.39±0.50 & 83.34±0.50 & 81.34±0.69 & 47.59±0.59 & 41.56±0.65 & 0.762±0.007 & 69.78±0.26\\


+ $\mathcal{P}ow$Mix & 83.17±0.80 & 81.45±0.65 & 83.43±0.46 & 81.41±0.47 & 48.16±0.31 & 42.03±0.55 & 0.757±0.010 & 69.94±0.36 \\

+ MS-Mix (ours) & \textbf{83.94}±0.52 & \textbf{82.21}±0.63 & \textbf{83.81}±0.32 & \textbf{82.14}±0.37 & \textbf{48.59}±0.56 & \textbf{42.36}±0.31 & \textbf{0.746}±0.005 & \textbf{70.51}±0.19  \\
\hline


\textbf{ALMT} & 83.48±0.80 & 82.06±0.61 & 83.39±0.78 & 81.94±0.59 & 51.32±0.75 & 45.46±0.86 & 0.733±0.003 & 71.28±0.23 \\
 
+ Manifold Mix & \textbf{84.16}±0.76 & 82.58±0.82 & 83.58±0.50 & 82.60±0.77 & 51.51±0.93 & 46.04±1.39  & 0.734±0.008 & 71.74±0.15 \\

+ MultiMix & 83.53±0.71 & 82.05±0.38 & 83.31±1.03 & 81.89±0.61  & 48.92±0.74 & 44.21±0.65 & 0.747±0.005 &  70.65±0.38 \\

+ AdvMixUp & 83.95±0.23 & 81.94±0.35 & 83.61±0.67  & 81.95±0.68 & 51.44±1.20  & 46.89±0.59 & 0.726±0.004 & 71.63±0.21  \\


+ $\mathcal{P}ow$Mix & 83.38±0.50  & 82.34±0.96 & 83.26±0.84 & 82.33±0.93 & 51.67±0.53 & 45.72±0.64 & 0.730±0.007 & 71.45±0.25 \\

+ MS-Mix (ours) & 84.02±0.51 & \textbf{83.21}±0.60 & \textbf{83.93}±0.33 & \textbf{83.02}±0.28 & \textbf{52.33}±0.65 & \textbf{46.98}±0.37  & \textbf{0.721}±0.004 &  \textbf{72.25}±0.32  \\
\hline


\textbf{GLoMo} & 83.81±0.27 & 82.19±0.32 & 83.73±0.28 & 82.13±0.33 & 52.59±0.71 & 47.10±0.94 & 0.743±0.010 & 71.92±0.22 \\
 
+ Manifold Mix & 84.78±0.43 & 82.95±0.62 & 84.80±0.23 & 83.00±0.70 & 52.89±0.96  & 49.45±0.99 & 0.731±0.006 & 72.98±0.18 \\

+ MultiMix & 84.94±0.56 & 83.23±0.58 & 85.02±0.49 & 83.32±0.58 & 53.01±0.47 & 50.46±0.93 & 0.728±0.004 & 73.33±0.22 \\

+ AdvMixUp & 84.27±0.73  & 83.24±0.65  &  84.39±0.59  & 83.25±0.65  & 53.27±0.49 & 51.09±0.68 & 0.730±0.004  & 73.25±0.36 \\

+ $\mathcal{P}ow$Mix & 84.81±0.29 & 83.36±0.19 & 84.84±0.36 & \textbf{83.48}±0.35 & 53.14±0.75 & 49.56±2.06 & 0.730±0.007 & 73.20±0.31 \\

+ MS-Mix (ours) & \textbf{85.21}±0.63 & \textbf{83.46}±0.19 & \textbf{85.23}±0.56 & 83.43±0.32 & \textbf{54.64}±0.96 & \textbf{52.31}±0.59 & \textbf{0.726}±0.005 & \textbf{74.04}±0.47  \\

\bottomrule
\end{tabular}
}
\label{table_mosi}
\end{table*}

\subsection{Verification Performance of MS-Mix}
Experiments were conducted on six backbones, representing both classical \cite{TFN2017, LMF2018, Mult2019, MISA2020} and state-of-the-art approaches \cite{ALMT2023,glomo2024}. For each model, we compared against classical Mixup variants, including Manifold Mix \cite{manifoldmix} and Multimix \cite{multimix}, as well as a recently proposed mixup-based method in MSA, $\mathcal{P}ow$Mix \cite{powmix}, and the adversarial-based Mixup method, AdvMixUp \cite{fu2025advmixup}. 
As shown in Tables \ref{table_mosi} and \ref{table_mosei}, MS-Mix achieves top performance on the comprehensive \textit{Avg.} metric across every backbone, ranking first or second across all metrics compared with the baseline and other methods. 

\begin{table*}[!ht]
\centering
\caption{Results of Various Approaches on the \textbf{MOSEI} Dataset (Mean ± std over 5 Independent Runs).
}

\resizebox{0.9\linewidth}{!}{

\begin{tabular}{c | c c c c  c c c  | c }
\toprule
\textbf{MODEL}  & w-ACC$_2$(\%)$\uparrow$ & ACC$_2$(\%)$\uparrow$ & w-F1(\%)$\uparrow$ & F1(\%)$\uparrow$ & ACC$_5$(\%)$\uparrow$ & ACC$_7$(\%)$\uparrow$ & MAE$\downarrow$ & Avg.(\%)$\uparrow$  \\
\hline
\textbf{TFN}$^\dag$ & 81.89 & 78.50 & 81.74 & 78.96 & 53.10 & 51.60 & 0.573 & 70.97  \\

+ Manifold Mix & \textbf{83.41}±0.73 & 81.42±0.48 & \textbf{83.42}±0.61 & 81.51±0.65 &  52.87±0.65 &  51.25±0.72 & 0.571±0.003 & 72.31±0.19 \\

+ MultiMix & 82.73±0.50 & 80.61±0.63 &  82.46±0.64 & 80.46±0.98 & 53.17±0.74 &  51.74±0.64 & 0.569±0.003 &  71.86±0.32 \\

+ AdvMixUp &  82.85±0.51  &  80.77±0.71 &   82.96±0.65 & 80.98±0.54  &  53.32±0.63  & \textbf{52.20}±0.69  &  0.566±0.002 & 72.18±0.16  \\

+ $\mathcal{P}ow$Mix & 83.36±1.00 & 81.25±0.65 &  83.37±0.72 & 81.51±0.53 & 53.08±0.58 & 51.64±0.55 &  0.564±0.004 & 72.37±0.56 \\

+ MS-Mix (ours) &  \textbf{83.41}±0.39 &  \textbf{82.16}±0.26 & 83.30±0.49 &  \textbf{82.07}±0.31 & \textbf{53.87}±0.44 & \textbf{52.20}±0.29 & \textbf{0.559}±0.002  & \textbf{72.84}±0.17   \\
\hline
\textbf{LMF}$^\dag$ & 83.48 & 80.54 & 83.36 & 80.94 & 52.99  & 51.59  & 0.576 & 72.15 \\
 
+ Manifold Mix & 83.21±0.64 & 78.91±1.49 & 83.39±0.68 &  79.01±1.46 & 53.40±0.83 & 52.04±0.76  & 0.572±0.008 &  71.66±0.44 \\

+ MultiMix & 84.17±0.71 &  80.73±0.80 & 84.29±0.58 &  80.80±0.83 & 53.72±0.88 & 52.72±0.96 & 0.564±0.002 & 72.74±0.36 \\

+ AdvMixUp &  84.12±0.55  & 81.72±0.81  & 84.12±0.63 & 81.79±0.75 & 53.47±1.04 & 52.14±0.77 & 0.565±0.008 & 72.90±0.38 \\

+ $\mathcal{P}ow$Mix & 83.95±0.31 & 80.97±0.59 & 84.02±0.30 & 81.27±0.53 & 54.37±0.59 & \textbf{52.73}±0.74  & 0.560±0.006 & 72.88±0.30 \\

+ MS-Mix (ours) & \textbf{84.53}±0.54 & \textbf{82.99}±0.48 &  \textbf{84.60}±0.40 & \textbf{83.04}±0.38 & 54.45±0.43 & 52.65±0.47  & \textbf{0.556}±0.004 & \textbf{73.71}±0.27 \\
\hline
\textbf{MuIT}$^\dag$ & 84.63 & 81.15 & 84.52 & 81.56 & 54.51 & 52.84 & 0.559  & 73.20 \\
 
+ Manifold Mix &  84.39±0.27 &  81.31±1.19 & 84.32±0.25 & 81.55±1.01 &  53.94±0.48 &  52.39±0.48 &  0.562±0.002 &  72.98±0.58 \\

+ MultiMix & 83.99±0.36  &  81.45±0.68  &  84.03±0.53  & 81.75±0.61  & 54.22±0.52  &  52.46±0.49 &  0.565±0.006   & 72.98±0.19\\

+ AdvMixUp & 84.31±0.50 & 81.17±0.17 &  84.29±0.71 & 80.94±0.74  & 54.30±0.44 & 51.98±0.17 & 0.570±0.003  &  72.83±0.33 \\


+ $\mathcal{P}ow$Mix &  84.71±0.62 & 81.74±0.75 & 84.65±0.90 & 81.85±0.35 & 54.37±0.51 & 52.68±0.47 & 0.559±0.003 & 73.33±0.23  \\

+ MS-Mix (ours) & \textbf{84.98}±0.34 & \textbf{82.08}±0.37 & \textbf{84.87}±0.73 & \textbf{82.19}±0.53 & \textbf{55.00}±0.39 & \textbf{53.04}±0.28 & \textbf{0.555}±0.003 & \textbf{73.69}±0.27 \\
\hline
\textbf{MISA}$^\dag$ & 84.67 & 80.67  & 84.66  & 81.12  & 53.63  & 52.05 & 0.558  &  72.80\\
 
+ Manifold Mix & 84.51±0.81  & 80.24±2.22  & 84.57±0.73  &  81.18±1.99 & 54.49±0.55  & 52.83±0.69  &  0.548±0.006  & 72.90±0.90 \\

+ MultiMix & 84.23±0.53 & 81.31±0.47  &   84.20±0.68 & 81.26±0.33 &  54.07±0.47 &  52.66±0.38 &  0.552±0.002  &   72.95±0.33 \\

+ AdvMixUp & 84.65±0.58  & 81.69±0.49  & 84.55±0.93 & 81.64±0.64 & 54.79±0.57 & 52.69±0.35 & 0.544±0.009  & 73.34±0.19  \\


+ $\mathcal{P}ow$Mix & 84.49±0.49 & 82.03±0.29 & 84.19±0.38 & 81.95±0.19 & 54.79±0.46  & 52.20±0.62 & 0.547±0.003  & 73.27±0.18 \\

+ MS-Mix (ours) & \textbf{84.84}±0.31  & \textbf{83.15}±0.36 & \textbf{84.72}±0.33 & \textbf{83.08}±0.51 & \textbf{55.04}±0.34 & \textbf{52.95}±0.43 & \textbf{0.542}±0.004 &  \textbf{73.96}±0.28  \\
\hline

\textbf{ALMT} & 85.15±0.80 & 81.35±0.56 & 85.04±0.48 & 81.53±0.50 & 54.11±0.29 & 52.73±0.32 & 0.550±0.003 & 73.32±0.14 \\
 
+ Manifold Mix & 85.13±1.46 & 82.16±1.06 & 85.08±1.33 & 81.65±2.73 & 54.33±0.83 & 52.65±1.00 & 0.543±0.005 & 73.33±1.19 \\

+ MultiMix & 85.10±1.02 & 82.00±0.73 & 85.22±0.60 & 81.96±0.78 & 54.70±0.74 & 52.70±0.70 & 0.538±0.007 & 73.61±0.50 \\

+ AdvMixUp & 85.22±0.56   & 81.99±0.32  &  85.05±0.64  & 82.03±0.42  & 54.46±0.54 & 52.87±0.30  & 0.544±0.008  &  73.60±0.32 \\


+ $\mathcal{P}ow$Mix & \textbf{85.55}±0.46 & 82.37±0.45 & 85.48±0.51 & 82.35±0.26 & 54.58±0.69 & 53.13±0.45 & 0.539±0.004 & 73.91±0.23    \\

+ MS-Mix (ours) & 85.41±0.62 &  \textbf{82.90}±0.66 & \textbf{85.62}±0.55 &  \textbf{83.08}±0.56 &  \textbf{55.17}±0.21 &  \textbf{53.55}±0.36 & \textbf{0.536}±0.003 &  \textbf{74.29}±0.30 \\
\hline
\textbf{GLoMo} & 84.93±0.80 & 82.71±0.56 & 85.28±0.50 & 82.95±0.77 & 53.80±0.48 & 52.67±0.60 & 0.542±0.004 & 73.72±0.52 \\
 
+ Manifold Mix & 84.88±0.90 & 83.65±0.57 & 85.13±0.72 & 83.28±0.50 & \textbf{56.08}±0.84 & \textbf{53.37}±1.53  & 0.535±0.011 & 74.40±0.48 \\

+ MultiMix & 85.13±0.39 & 83.29±0.67 & \textbf{85.25}±0.39 & 82.92±0.81 & 55.05±0.28 & 52.93±0.33 & 0.535±0.005 & 74.09±0.27 \\

+ AdvMixUp &  85.24±0.40  & 83.20±0.39  & 84.92±0.42 & 82.98±0.35 & 55.00±0.39  & 53.04±0.36 &  0.537±0.004 & 74.06±0.22 \\

+ $\mathcal{P}ow$Mix & 85.08±0.20 & 83.19±0.23 & 85.12±0.26 & 83.38±0.48 & 55.08±0.29 & 53.07±0.27 & 0.533±0.003 & 74.15±0.07 \\

+ MS-Mix (ours) & \textbf{85.32}±0.26 & \textbf{84.12}±0.31 & \textbf{85.25}±0.35 & \textbf{84.17}±0.38 & 55.84±0.22 & 53.15±0.30 & \textbf{0.53}1±0.003 &  \textbf{74.64}±0.10  \\

\bottomrule
\end{tabular}
}
\label{table_mosei}
\end{table*}

The consistent outperformance of MS-Mix across diverse backbones and datasets stems from its core design principles, which directly address key limitations in existing mixup-based augmentation methods:
First, by ensuring semantic and label consistency, SASS significantly enhances the quality of augmented data. In contrast, the comparative methods mix random sample pairs regardless of emotional coherence, easily yielding noisy labels. 
Second, the SIG mixing module adaptively determines modality-specific mixing ratios based on emotional salience, rather than using random or fixed values \cite{manifoldmix, multimix}, or leveraging modality importance without emotional context \cite{powmix, fu2025advmixup}. This allows MS-Mix to dynamically weight each modality's contribution, leading to more discriminative mixed features.
Third, the SAL serves as a regularization term that aligns predicted sentiment distributions with ground-truth labels, enhancing prediction consistency across modalities and providing additional supervisory signal. While the comparative methods lack explicit constraints on the alignment of emotion-level distributions.

\subsection{Performance Visualization}
To further demonstrate the effectiveness of MS-Mix, we conduct both t-SNE~\cite{maaten2008tsne} visualizations (Fig. \ref{fig_sne}) and loss landscape visualizations (Fig. \ref{fig_loss}). 
Specifically, we employ t-SNE to project the distributions of textual features for both the original and mixed features extracted from the MOSEI dataset, using the MISA model. The results reveal that features generated by MS-Mix exhibit significantly clearer decision boundaries and more distinct cluster separation than those produced by $\mathcal{P}$owMix. 

The generalization ability of neural networks is closely related to the geometry of their loss landscapes, where convergence to flat minima is generally associated with superior robustness and generalization performance~\cite{foret2020sharpness}. To visualize this, we plot the 2D loss surface of the ALMT model on MOSEI by perturbing the parameters along two random, mutually orthogonal directions, generated via Gaussian sampling and Gram-Schmidt orthogonalization.
As shown in Fig. \ref{fig_loss}, introducing Manifold Mix and other variants resulted in some landscape smoothing. However, pronounced peaks and high local curvature remained, indicating persistent sensitivity to parameter perturbations. In contrast, MS-Mix yields a notably smoother, flatter landscape, suggesting reduced gradient conflict during training and convergence toward more robust minima.

\begin{figure*}[htbp]
\centerline{\includegraphics[scale=0.80]{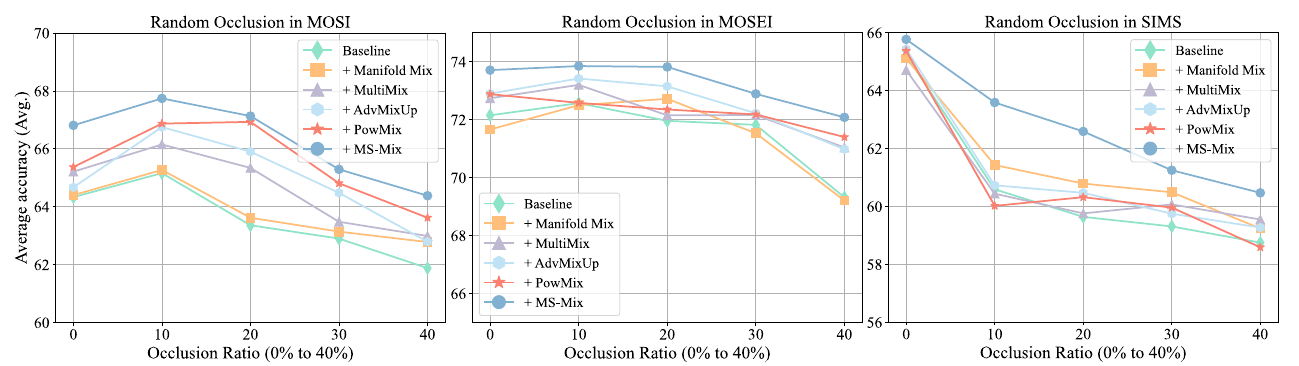}}
    \caption{The performance of different mixup-based methods at different occlusion ratios on three datasets.
    }
    \vspace{-5pt}
    \label{fig5}
\end{figure*}
\begin{table*}[!ht]
\centering
\caption{Results of the ablation experiments using the \textbf{LMF} method on the \textbf{MOSI} database.}

\setlength{\tabcolsep}{0.6mm}
\resizebox{0.66\linewidth}{!}{

\begin{tabular}{c c  c | c c c  c c c c | c }
\toprule
 SASS & SIG & SAL  & w-ACC$_2$(\%)$\uparrow$ & ACC$_2$(\%)$\uparrow$ & w-F1(\%)$\uparrow$ & F1(\%)$\uparrow$ & ACC$_5$(\%)$\uparrow$ & ACC$_7$(\%)$\uparrow$ &  MAE$\downarrow$ & Avg.(\%)$\uparrow$ \\
\hline
  &  &  &  79.23 & 77.85 & 79.34 & 77.95 & 37.84 & 34.17 & 0.943 & 64.40 \\

 \checkmark &  &  & 79.98 & 78.41  & 80.08 & 78.60 & 38.86 & 35.26  & 0.921 & 65.20  \\

  & \checkmark &  & 80.94 & 78.91 & 80.93 & 79.00 & 40.48 & 35.25 & 0.925 & 65.92  \\

  & \checkmark & \checkmark & 81.48 & 79.71 & 81.58 & 79.38 & 41.54 & 35.37  &  0.949  &  66.51 \\
 
 \checkmark& \checkmark &  &  81.41 & 78.03 & 81.41 & 79.61  & 40.48 & 36.00 & 0.919 & 66.16  \\

 \checkmark & \checkmark & \checkmark & \textbf{82.02} & \textbf{79.16} & \textbf{82.01} & \textbf{79.27} &  \textbf{41.90} &  \textbf{36.52} &  \textbf{0.879} & \textbf{66.81} \\



\bottomrule
\end{tabular}
}
\label{table_ablation}
\end{table*}

\subsection{Ablation Analysis}
To rigorously quantify the contribution of each component, we conduct ablation studies on the LMF backbone \cite{LMF2018} using the MOSI \cite{MOSI}. We incrementally activate SASS, SIG, and SAL (SAL depends on SIG’s intensity predictor and cannot be isolated). Results in Table~\ref{table_ablation}. SASS delivers a clear gain by selecting samples before mixing. This eliminates contradictory emotion pairs, and the effect is confirmed by sharper decision boundaries in t-SNE visualization (Fig. \ref{fig_sne}). SIG further improves \textit{Avg.} via computing modality-specific $\lambda^m$ conditioned on emotional intensity $I^m$. This dynamic weighting ensures high-salience modalities dominate the mix. SAL aligns predicted sentiment distributions $\mathbf{P}^m$ with ground-truth $\mathbf{P}^L$ across modalities, imposing cross-modal consistency constraints during training and thereby enabling the model to achieve optimal performance.

These gains, combined with the smoother loss landscape observed for MS-Mix (Fig. \ref{fig_loss}), demonstrate that our sentiment-guided design choices are not incremental tweaks but principled solutions tailored to the unique challenges of MSA.

\subsection{Occlusion Experiment}
In the random occlusion experiment, we compared the performance of our method against baseline approaches \cite{manifoldmix,multimix,powmix} under both clean and noisy data conditions across MOSI \cite{MOSI}, MOSEI \cite{MOSEI}, and SIMS \cite{sims}, which is a commonly used Chinese MSA dataset (refer to Appendix A.2 for details). Specifically, we randomly masked out 0\% to 40\% of the input data during training of the LMF \cite{LMF2018} model to investigate the effect of mixup-based methods on model robustness.

As shown in Fig. \ref{fig5}, the experimental results indicate that introducing noise affects model performance across all datasets. Despite these variations, MS-Mix consistently achieves the highest average accuracy across all evaluated conditions, demonstrating its advantage in optimizing the model to handle noisy data.

\section{CONCLUSION}
In this paper, we propose MS-Mix, a novel, adaptive data augmentation framework designed specifically for MSA. By integrating sentiment-aware sample selection, sentiment-guided mixing, and cross-modal alignment regularization. MS-Mix effectively mitigates semantic inconsistency and label noise inherent in traditional mixup methods.
An empirical evaluation across public benchmark datasets and six diverse backbone architectures shows that MS-Mix achieves state-of-the-art performance.
Future work will extend MS-Mix to other multimodal tasks such as mental health monitoring and human-computer interaction, and explore self-supervised strategies to further alleviate annotation costs.
Finally, from an efficiency perspective, we will focus on developing lightweight variants of the sample selection and fusion mechanisms to reduce computational costs and support deployment on resource-constrained devices.

\begin{acks}
To Robert, for the bagels and explaining CMYK and color spaces.
\end{acks}

\bibliographystyle{ACM-Reference-Format}
\bibliography{reference}

@String(ICCV  = {Int. Conf. Comput. Vis.})

@String(ECCV  = {Eur. Conf. Comput. Vis.})

@String(NeurIPS = {Adv. Neural Inform. Process. Syst.})

@String(ICML  = {Int. Conf. Mach. Learn.})

@String(ICLR  = {Int. Conf. Learn. Represent.})

@String(AAAI  = {AAAI})

@String(ICCV  = {ICCV})

@String(ECCV  = {ECCV})

@String(NeurIPS = {NeurIPS})

@String(ICML  = {ICML})

@String(ICLR  = {ICLR})

@article{MOSI,
  title={Multimodal sentiment intensity analysis in videos: Facial gestures and verbal messages},
  author={Zadeh, Amir and Zellers, Rowan and Pincus, Eli and Morency, Louis-Philippe},
  journal={IEEE Intelligent Systems},
  volume={31},
  number={6},
  pages={82--88},
  year={2016},
  publisher={IEEE}
}

@inproceedings{MOSEI,
    author = {Zadeh, AmirAli Bagher and Liang, Paul Pu and Poria, Soujanya and Cambria, Erik and Morency, Louis-Philippe},
    title = {Multimodal language analysis in the wild: CMU-mosei dataset and interpretable dynamic fusion graph},
    booktitle = {Proceedings of the Association for computational linguistics (ACL)},
    volume={1},
    pages={2236--2246},
    year = {2018}
}

@inproceedings{sims,
  title={Ch-sims: A chinese multimodal sentiment analysis dataset with fine-grained annotation of modality},
  author={Yu, Wenmeng and Xu, Hua and Meng, Fanyang and Zhu, Yilin and Ma, Yixiao and Wu, Jiele and Zou, Jiyun and Yang, Kaicheng},
  booktitle={Proceedings of the Association for Computational Linguistics (ACL)},
  pages={3718--3727},
  year={2020}
}

@inproceedings{multimix,
  title={Embedding space interpolation beyond mini-batch, beyond pairs and beyond examples},
  author={Venkataramanan, Shashanka and Kijak, Ewa and Avrithis, Yannis and others},
  booktitle={Advances in Neural Information Processing Systems (NeurIPS)},
  volume={36},
  pages={61923--61935},
  year={2023},
  publisher={Curran Associates, Inc.}
}

@inproceedings{manifoldmix,
  title={Manifold mixup: Better representations by interpolating hidden states},
  author={Verma, Vikas and Lamb, Alex and Beckham, Christopher and Najafi, Amir and Mitliagkas, Ioannis and Lopez-Paz, David and Bengio, Yoshua},
  booktitle={International Conference on Machine Learning (ICML)},
  pages={6438--6447},
  year={2019},
  publisher={PMLR}
}

@inproceedings{vanilla_mixup,
  title={mixup: Beyond Empirical Risk Minimization},
  author={Zhang, Hongyi and Cisse, Moustapha and Dauphin, Yann N and Lopez-Paz, David},
  booktitle={International Conference on Learning Representations (ICLR)},
  year={2018},
  publisher={OpenReview.net}
}

@inproceedings{puzzlemix,
  title={Puzzle mix: Exploiting saliency and local statistics for optimal mixup},
  author={Kim, Jang-Hyun and Choo, Wonho and Song, Hyun Oh},
  booktitle={International Conference on Machine Learning (ICML)},
  pages={5275--5285},
  year={2020},
  publisher={PMLR}
}

@inproceedings{automix,
  title={Automix: Unveiling the power of mixup for stronger classifiers},
  author={Liu, Zicheng and Li, Siyuan and Wu, Di and Liu, Zihan and Chen, Zhiyuan and Wu, Lirong and Li, Stan Z},
  booktitle={European Conference on Computer Vision (ECCV)},
  pages={441--458},
  year={2022},
  publisher={Springer}
}

@article{KLloss,
  title={R{\'e}nyi divergence and Kullback-Leibler divergence},
  author={Van Erven, Tim and Harremos, Peter},
  journal={IEEE Transactions on Information Theory},
  volume={60},
  number={7},
  pages={3797--3820},
  year={2014},
  publisher={IEEE}
}

@inproceedings{attention,
  title={Attention is all you need},
  author={Vaswani, Ashish and Shazeer, Noam and Parmar, Niki and Uszkoreit, Jakob and Jones, Llion and Gomez, Aidan N and Kaiser, {\L}ukasz and Polosukhin, Illia},
  booktitle={Advances in Neural Information Processing Systems (NeurIPS)},
  volume={30},
  year={2017},
  publisher={Curran Associates, Inc.}
}

@article{borovits2026addressing,
  title={Addressing data scarcity with synthetic data: a secure and gdpr-compliant cloud-based platform},
  author={Borovits, Nemania and Bardelloni, Gianluigi and Hashemi, Hossein and Tulsiani, Masoom and Tamburri, Damian Andrew and van den Heuvel, Willem-Jan},
  journal={ACM Transactions on Software Engineering and Methodology},
  volume={35},
  number={3},
  pages={1--50},
  year={2026},
  publisher={ACM New York, NY}
}

@article{li2019survey,
  title={A survey on sentiment analysis and opinion mining for social multimedia},
  author={Li, Zuhe and Fan, Yangyu and Jiang, Bin and Lei, Tao and Liu, Weihua},
  journal={Multimedia Tools and Applications},
  volume={78},
  number={6},
  pages={6939--6967},
  year={2019},
  publisher={Springer}
}

@article{wulamu2025enhanced,
  title={Enhanced multi-modal emotion recognition using the feature level fusion},
  author={Wulamu, Aziguli and Wu, Yuheng and Liu, Xin and Zhang, Yao and Xu, Jinghan and Zhang, Yang},
  journal={Engineering Applications of Artificial Intelligence},
  volume={162},
  pages={112447},
  year={2025},
  publisher={Elsevier}
}

@article{powmix,
  title={$\mathcal{P}$owMix: A Versatile Regularizer for Multimodal Sentiment Analysis},
  author={Georgiou, Efthymios and Avrithis, Yannis and Potamianos, Alexandros},
  journal={IEEE/ACM Transactions on Audio, Speech, and Language Processing},
  volume={32},
  pages={5010--5023},
  year={2024},
  publisher={IEEE}
}

@inproceedings{Mult2019,
  title={Multimodal transformer for unaligned multimodal language sequences},
  author={Tsai, Yao-Hung Hubert and Bai, Shaojie and Liang, Paul Pu and Kolter, J Zico and Morency, Louis-Philippe and Salakhutdinov, Ruslan},
  booktitle={Proceedings of the Association for Computational Linguistics (ACL).},
  volume={2019},
  pages={6558},
  year={2019}
}

@inproceedings{foret2020sharpness,
  title={Sharpness-aware minimization for efficiently improving generalization},
  author={Foret, Pierre and Kleiner, Ariel and Mobahi, Hossein and Neyshabur, Behnam},
   booktitle={International Conference on Learning Representations (ICLR)},
  year={2021}
}

@inproceedings{MSENA2022,
  title={M-SENA: An Integrated Platform for Multimodal Sentiment Analysis},
  author={Mao, Huisheng and Yuan, Ziqi and Xu, Hua and Yu, Wenmeng and Liu, Yihe and Gao, Kai},
  booktitle={Proceedings of the 60th Annual Meeting of the Association for Computational Linguistics: System Demonstrations},
  pages={204--213},
  year={2022}
}

@article{2014dropout,
  title={Dropout: a simple way to prevent neural networks from overfitting},
  author={Srivastava, Nitish and Hinton, Geoffrey and Krizhevsky, Alex and Sutskever, Ilya and Salakhutdinov, Ruslan},
  journal={The journal of machine learning research},
  volume={},
  number={},
  pages={1929--1958},
  year={2014},
  publisher={JMLR. org}
}

@inproceedings{2019cutmix,
  title={Cutmix: Regularization strategy to train strong classifiers with localizable features},
  author={Yun, Sangdoo and Han, Dongyoon and Oh, Seong Joon and Chun, Sanghyuk and Choe, Junsuk and Yoo, Youngjoon},
  booktitle={Proceedings of the IEEE/CVF international conference on computer vision (ICCV)},
  pages={6023--6032},
  year={2019}
}

@inproceedings{2021saliencymix,
  title={SALIENCYMIX: A SALIENCY GUIDED DATA AUGMENTATION STRATEGY FOR BETTER REGULARIZATION},
  author={Uddin, AFM Shahab and Monira, Mst Sirazam and Shin, Wheemyung and Chung, Tae Choong and Bae, Sung Ho},
  booktitle={the 9th International Conference on Learning Representations (ICLR) },
  year={2021},
  publisher={OpenReview.net}
}

@inproceedings{ba2016layer,
  title={Layer Normalization},
  author={Ba, Jimmy Lei and Kiros, Jamie Ryan and Hinton, Geoffrey E},
  booktitle={Advances in Neural Information Processing Systems (NeurIPS)},
  volume={29},
  pages={601--610},
  year={2016},
  publisher={Curran Associates, Inc.}
}

@inproceedings{TFN2017,
  title={Tensor Fusion Network for Multimodal Sentiment Analysis},
  author={Zadeh, Amir and Chen, Minghai and Poria, Soujanya and Cambria, Erik and Morency, Louis-Philippe},
  booktitle={Proceedings of the Conference on Empirical Methods in Natural Language Processing (EMNLP)},
  pages={1103--1114},
  year={2017}
}

@inproceedings{LMF2018,
  title={Efficient Low-rank Multimodal Fusion with Modality-Specific Factors},
  author={Liu, Zhun and Shen, Ying},
  booktitle={Proceedings of the Association for Computational Linguistics (ACL)},
  volume={1},
  pages={2247–2256},
  year={2018}
}

@inproceedings{MISA2020,
  title={Misa: Modality-invariant and-specific representations for multimodal sentiment analysis},
  author={Hazarika, Devamanyu and Zimmermann, Roger and Poria, Soujanya},
  booktitle={Proceedings of the 28th ACM international conference on multimedia},
  pages={1122--1131},
  year={2020}
}

@article{weldy2026simulated,
  title={Simulated soundscapes and transfer learning boost the performance of acoustic classifiers under data scarcity},
  author={Weldy, Matthew J and Lesmeister, Damon B and Denton, Tom and Duarte, Adam and Vernasco, Ben J and Gasc, Amandine and Rowe, Jennifer C and Adams, Michael J and Betts, Matthew G},
  journal={Methods in Ecology and Evolution},
  volume={17},
  number={2},
  pages={322--338},
  year={2026},
  publisher={Wiley Online Library}
}

@article{huang2026ai,
  title={When AI Becomes a Friend: The “Emotional” and “Rational” Mechanism of Problematic Use in Generative AI Chatbot Interactions},
  author={Huang, Hanyun and Shi, Lihong and Pei, Xin},
  journal={International Journal of Human--Computer Interaction},
  volume={42},
  number={6},
  pages={4006--4024},
  year={2026},
  publisher={Taylor \& Francis}
}

@inproceedings{ALMT2023,
  title={Learning Language-guided Adaptive Hyper-modality Representation for Multimodal Sentiment Analysis},
  author={Zhang, Haoyu and Wang, Yu and Yin, Guanghao and Liu, Kejun and Liu, Yuanyuan and Yu, Tianshu},
  booktitle={The Conference on Empirical Methods in Natural Language Processing (EMNLP)},
  page={756–-767},
  year={2023}
}

@inproceedings{glomo2024,
  title={GLoMo: Global-local modal fusion for multimodal sentiment analysis},
  author={Zhuang, Yan and Zhang, Yanru and Hu, Zheng and Zhang, Xiaoyue and Deng, Jiawen and Ren, Fuji},
  booktitle={Proceedings of the 32nd ACM International Conference on Multimedia (ACM MM)},
  pages={1800--1809},
  year={2024}
}

@inproceedings{2018memoryzadeh,
  title={Memory fusion network for multi-view sequential learning},
  author={Zadeh, Amir and Liang, Paul Pu and Mazumder, Navonil and Poria, Soujanya and Cambria, Erik and Morency, Louis-Philippe},
  booktitle={Proceedings of the AAAI conference on artificial intelligence},
  volume={32},
  number={1},
  year={2018}
}

@inproceedings{2021attentionliang,
  title={Attention is not enough: Mitigating the distribution discrepancy in asynchronous multimodal sequence fusion},
  author={Liang, Tao and Lin, Guosheng and Feng, Lei and Zhang, Yan and Lv, Fengmao},
  booktitle={Proceedings of the IEEE/CVF International Conference on Computer Vision},
  pages={8148--8156},
  year={2021}
}

@inproceedings{2021progressivelv,
  title={Progressive modality reinforcement for human multimodal emotion recognition from unaligned multimodal sequences},
  author={Lv, Fengmao and Chen, Xiang and Huang, Yanyong and Duan, Lixin and Lin, Guosheng},
  booktitle={Proceedings of the IEEE/CVF conference on computer vision and pattern recognition},
  pages={2554--2562},
  year={2021}
}

@inproceedings{li2023decoupled,
  title={Decoupled multimodal distilling for emotion recognition},
  author={Li, Yong and Wang, Yuanzhi and Cui, Zhen},
  booktitle={Proceedings of the IEEE/CVF conference on computer vision and pattern recognition},
  pages={6631--6640},
  year={2023}
}

@inproceedings{yang2022dis,
  title={Disentangled representation learning for multimodal emotion recognition},
  author={Yang, Dingkang and Huang, Shuai and Kuang, Haopeng and Du, Yangtao and Zhang, Lihua},
  booktitle={Proceedings of the 30th ACM international conference on multimedia},
  pages={1642--1651},
  year={2022}
}

@inproceedings{han2021improving,
  title={Improving Multimodal Fusion with Hierarchical Mutual Information Maximization for Multimodal Sentiment Analysis},
  author={Han, Wei and Chen, Hui and Poria, Soujanya},
  booktitle={Proceedings of the Conference on Empirical Methods in Natural Language Processing (EMNLP)},
  pages={9180--9192},
  year={2021}
}

@article{lin2023multi,
  title={Multi-task momentum distillation for multimodal sentiment analysis},
  author={Lin, Ronghao and Hu, Haifeng},
  journal={IEEE Transactions on Affective Computing},
  volume={15},
  number={2},
  pages={549--565},
  year={2023},
  publisher={IEEE}
}

@inproceedings{SUMIX2024,
  title={Sumix: Mixup with semantic and uncertain information},
  author={Qin, Huafeng and Jin, Xin and Zhu, Hongyu and Liao, Hongchao and El-Yacoubi, Moun{\^\i}m A and Gao, Xinbo},
  booktitle={European Conference on Computer Vision (ECCV)},
  pages={70--88},
  year={2024},
  publisher={Springer}
}

@article{jin2024survey,
  title={A survey on mixup augmentations and beyond},
  author={Jin, Xin and Zhu, Hongyu and Li, Siyuan and Wang, Zedong and Liu, Zicheng and Tian, Juanxi and Yu, Chang and Qin, Huafeng and Li, Stan Z},
  journal={arXiv preprint arXiv:2409.05202},
  year={2024}
}

@article{lin2024adapt,
  title={Adapt and explore: Multimodal mixup for representation learning},
  author={Lin, Ronghao and Hu, Haifeng},
  journal={Information Fusion},
  volume={105},
  pages={102216},
  year={2024},
  publisher={Elsevier}
}

@inproceedings{2022vlmixer,
  title={Vlmixer: Unpaired vision-language pre-training via cross-modal cutmix},
  author={Wang, Teng and Jiang, Wenhao and Lu, Zhichao and Zheng, Feng and Cheng, Ran and Yin, Chengguo and Luo, Ping},
  booktitle={International Conference on Machine Learning (ICML)},
  pages={22680--22690},
  year={2022},
  publisher={PMLR}
}

@article{xiao2025exploring,
  title={Exploring cognitive and aesthetic causality for multimodal aspect-based sentiment analysis},
  author={Xiao, Luwei and Mao, Rui and Zhao, Shuai and Lin, Qika and Jia, Yanhao and He, Liang and Cambria, Erik},
  journal={IEEE Transactions on Affective Computing},
  year={2025},
  pages={1--18},
  publisher={IEEE}
}

@article{maaten2008tsne,
  title={Visualizing data using t-SNE},
  author={Maaten, Laurens van der and Hinton, Geoffrey},
  journal={Journal of machine learning research},
  volume={9},
  number={Nov},
  pages={2579--2605},
  year={2008}
}

@article{fu2025advmixup,
  title={AdvMixUp: Adversarial MixUp Regularization for Deep Learning},
  author={Fu, Jun and Ji, Xianrui and Chen, Dexiong and Hu, Guosheng and Li, Shuang and Feng, Xiating},
  journal={IEEE Transactions on Neural Networks and Learning Systems},
  year={2025},
  publisher={IEEE}
}

@inproceedings{lv2025diffufuse,
  title={DiffuFuse: Diffusion-Driven Dual-Stream Fusion Framework for Multimodal Sentiment Analysis},
  author={Lv, Xiongjian and Wen, Yimin and Yu, Hang},
  booktitle={Proceedings of the 33rd ACM International Conference on Multimedia},
  pages={8458--8467},
  year={2025}
}




\end{document}